\documentclass[letterpaper]{article} 
\usepackage{aaai2027}  
\usepackage[hyphens]{url}  
\usepackage{graphicx} 
\usepackage{natbib}  
\usepackage{caption} 
\usepackage{algorithm}
\usepackage{algorithmic}

\usepackage{newfloat}
\usepackage{listings}
\DeclareCaptionStyle{ruled}{labelfont=normalfont,labelsep=colon,strut=off} 
\floatstyle{ruled}
\newfloat{listing}{tb}{lst}{}
\floatname{listing}{Listing}

\usepackage{booktabs}
\usepackage{amsmath}
\usepackage{amssymb}
\usepackage{pifont}
\usepackage{graphicx}
\usepackage{array}
\usepackage[most]{tcolorbox}
\usepackage{listings}
\usepackage{tabularx}
\usepackage{capt-of}

\newcommand{\method}{ArborMem}
\newcommand{\benchmark}{BranchMemEval}

\title{\method{}: Navigating Interaction States with Memory Forests}
\author{
    Zongwei Lv\textsuperscript{\rm 1}\thanks{The first four authors contributed equally; their ordering was randomized.},
    Yuemeng Xu\textsuperscript{\rm 1}\footnotemark[1],
    Yilun Yao\textsuperscript{\rm 1}\footnotemark[1],
    Dingsiyi\textsuperscript{\rm 1}\footnotemark[1],
    Xinyu Tan\textsuperscript{\rm 2},
    Yaoming Li\textsuperscript{\rm 1},
    Guangxiang Zhao\textsuperscript{\rm 3},
    Weihong Lin\textsuperscript{\rm 4},
    Lin Sun\textsuperscript{\rm 4},
    Xiangzheng Zhang\textsuperscript{\rm 4},
    Tong Yang\textsuperscript{\rm 1}\corresponding
}
\affiliations{
    \textsuperscript{\rm 1}Peking University\\
    \textsuperscript{\rm 2}Beijing University of Posts and Telecommunications\\
    \textsuperscript{\rm 3}Qiyuan Tech\\
    \textsuperscript{\rm 4}Qihoo 360
}

\begin{document}

\maketitle


\begin{abstract}
Large language models increasingly serve as persistent conversational assistants, requiring memory that preserves relevant experience and maintains continuity across interactions. Existing methods improve access to conversational history through long-context processing, selective retrieval, and structured memory organization. However, most systems treat memory access as retrieving relevant past information without first determining which prior interaction state the current turn resumes. This limitation becomes particularly important when conversations interleave multiple tasks, people, and plans that may be interrupted and later revisited.
We introduce \textbf{\method{}}, an online memory framework that represents a long-running conversation as a navigable forest of interaction states. Each branch preserves a locally coherent trajectory, while the forest maintains multiple trajectories that may later be resumed. 
For each new input, \method{} localizes the relevant state, restores its branch-local context, and augments it with reusable evidence retrieved across branches, preserving interaction continuity without conflating semantically related but structurally distinct trajectories.
Existing long-term memory benchmarks cover diverse memory and reasoning capabilities but do not explicitly isolate branch-structured challenges. We therefore introduce \textbf{\benchmark{}}, a controlled diagnostic benchmark for interleaved and resumable interaction trajectories. Experiments on LongMemEval, LoCoMo, BEAM 100K, and \benchmark{} show that \method{} outperforms the strongest baselines by 3.36--10.31 percentage points on the three established benchmarks and by 5.0 points on \benchmark{}.
Its advantage grows under constrained read budgets, while complete memory queries remain below half a second.
\end{abstract}
\section{Introduction}
\label{sec:intro}

Large language models increasingly serve as persistent conversational assistants rather than one-shot question-answering systems. 
As interactions extend over days or weeks, users expect assistants to remember prior discussions, track ongoing tasks, and resume earlier plans. Yet a language model has no durable state beyond the information available in its current context. Long-running conversational agents therefore require memory that preserves relevant experience and maintains continuity as users' needs evolve~\citep{maharana-etal-2024-evaluating}.

Existing methods improve access to conversational history in several ways. Long-context models expose more history directly, with recent frontier models supporting context windows on the order of one million tokens~\citep{geminiteam2024gemini15unlockingmultimodal}. However, context windows remain bounded, and making information visible does not ensure that it will be identified and used reliably~\citep{liu-etal-2024-lost}. Selective memory systems instead externalize
history as retrieved passages, summaries, episodic records, or extracted facts, as in Generative Agents~\citep{Park_2023}, MemoryBank~\citep{Zhong_Guo_Gao_Ye_Wang_2024}, and
Mem0~\citep{chhikara2025mem0buildingproductionreadyai}. Other methods organize memory through hierarchical stores, linked notes, trees, or temporal knowledge graphs, including MemGPT~\citep{packer2024memgptllmsoperatingsystems}, A-MEM~\citep{NEURIPS2025_19909c36}, MemTree~\citep{rezazadeh2025isolatedconversationshierarchicalschemas}, and Graphiti~\citep{rasmussen2025zeptemporalknowledgegraph}.

\begin{figure}[t]
\centering
\includegraphics[width=1.0\columnwidth]
{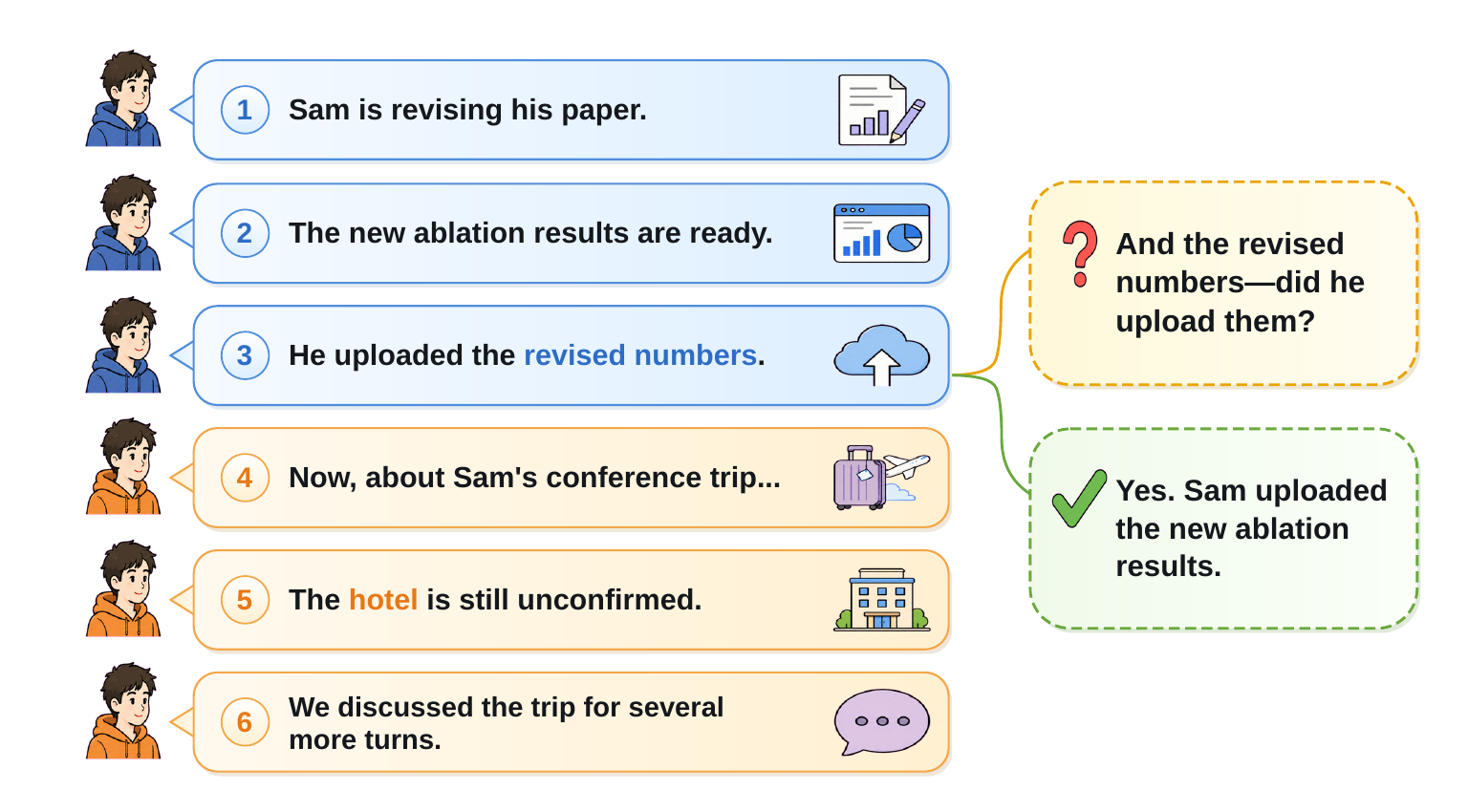}
\caption{A motivating example of interaction-state localization.
Although the most recent turns concern Sam's conference trip, the
underspecified follow-up resumes the earlier paper-revision
trajectory.}
\label{fig:case}
\end{figure}

Despite this progress, most systems treat memory access as retrieving relevant past information without first determining which prior interaction state the current turn resumes. 
Figure~\ref{fig:case} illustrates the distinction. The conversation interleaves two persistent trajectories concerning Sam's paper revision and conference trip.
Although the latest turns concern the trip, the underspecified follow-up returns to the earlier paper discussion. A relevance-based retriever may therefore mix the two threads, whereas correct interpretation requires first identifying the interaction context resumed by the current turn.
Such situations are common in persistent interaction, where assistants must manage multiple tasks, people, projects, and plans that may be interrupted and later resumed. Because these trajectories may involve overlapping entities or similar states while remaining distinct, long-term memory must preserve continuity across them rather than merely retain information.

We introduce \textbf{\method}, an online memory framework that represents a long-running conversation as a navigable forest of interaction states. Each branch preserves a locally coherent trajectory, while the forest maintains multiple trajectories that may later be resumed. For each new input, \method{} identifies the relevant branch, restores its local context, and supplements it with reusable evidence retrieved across branches. This design preserves interaction continuity while enabling information reuse without conflating semantically related but structurally distinct trajectories.


Existing long-term memory benchmarks broadly cover factual recall, temporal reasoning, knowledge updates, multi-hop reasoning, and very long conversations, but do not explicitly isolate branch-structured challenges such as topic interleaving, delayed trajectory resumption, and confusion between parallel agendas.
To address this gap, we further introduce \textbf{\benchmark{}}, a controlled diagnostic benchmark for branch-structured conversational memory.

We evaluate \method{} on LongMemEval, LoCoMo, BEAM 100K, and \benchmark{}.
\method{} outperforms the strongest baselines by 3.36 to 10.31 percentage points on the three established benchmarks and by 5.0 points on \benchmark{}. Its advantage grows under
constrained read budgets, indicating that state localization supports more effective evidence selection when only a small portion of memory can be accessed. Complete memory queries remain below half a second despite the additional routing and context assembly.

Our contributions are threefold.
First, we identify interaction-state localization as a key requirement for reliable long-term conversational memory: a system should determine which prior trajectory the current turn resumes before selecting
historical evidence.
Second, we propose \method{}, which organizes evolving conversational states as a navigable memory forest, restores coherent branch-local context, and retrieves reusable evidence across branches without conflating distinct interaction trajectories.
Third, we introduce \benchmark{}, a controlled diagnostic benchmark for branch-structured conversational memory, and demonstrate that \method{} consistently outperforms strong baselines on both established long-term memory benchmarks and our controlled evaluation.

\section{Related Work}

\subsection{Long-Term and Structured Agent Memory}

Long-context models expose more interaction history, but evaluations show that visibility does not guarantee reliable access and that performance remains sensitive to evidence position~\citep{bai-etal-2024-longbench,hsieh2024rulerwhatsrealcontext, liu-etal-2024-lost}.
Selective memory methods instead externalize history as turns, summaries, facts, or coherent segments~\citep{NEURIPS2020_6b493230,
pan2025secom,
tan-etal-2025-prospect}.

Agent memory systems further support retrieval, reflection, updating, forgetting, consolidation, and hierarchical management~\citep{Park_2023,Zhong_Guo_Gao_Ye_Wang_2024, chhikara2025mem0buildingproductionreadyai, packer2024memgptllmsoperatingsystems}. Recent systems further organize memories through linked notes, hierarchical trees, or temporal, event-centric, and entity-relation
graphs~\citep{NEURIPS2025_19909c36,
rezazadeh2025isolatedconversationshierarchicalschemas, zhang-etal-2025-bridging, huang-etal-2026-licomemory, rasmussen2025zeptemporalknowledgegraph}.

These approaches primarily determine what information to store, organize, or retrieve, with structures based on semantic association, hierarchy, events, or entity relations. In contrast, \method{} uses memory topology to represent continuity across interaction trajectories and retrieves reusable evidence separately, preventing semantically related but distinct states from being merged.

\subsection{Conversation Structure and Interaction-State Localization}

Conversation structure has been studied through topic segmentation, reply prediction, and conversation disentanglement, which recover threads or hierarchies from interleaved dialogue~\citep{arguello-rose-2006-topic, louis-cohen-2015-conversation, kummerfeld-etal-2019-large, yu-joty-2020-online}. These studies show that conversation cannot always be represented as a single chronological sequence, but primarily analyze fixed transcripts.

In contrast, \method{} maintains interaction structure as an evolving memory state and decides online whether each input continues an existing trajectory or begins a new one. Its interaction states capture not only topics or reply relations, but also the tasks, entities, goals, assumptions, and local context required for interpretation. 

Existing benchmarks cover factual, temporal, and multi-session memory, with EverMemBench further considering cross-topic interleaving and thread resumption~\citep{wu2025longmemevalbenchmarkingchatassistants, maharana-etal-2024-evaluating, tavakoli2026milliontokensbenchmarkingenhancing, hu2026evaluatinglonghorizonmemorymultiparty}.
However, they do not isolate branch localization, delayed resumption, or parallel-thread confusion as controlled diagnostic factors; \benchmark{} targets these capabilities.
\begin{figure*}[t]
\centering
\includegraphics[width=\textwidth]
{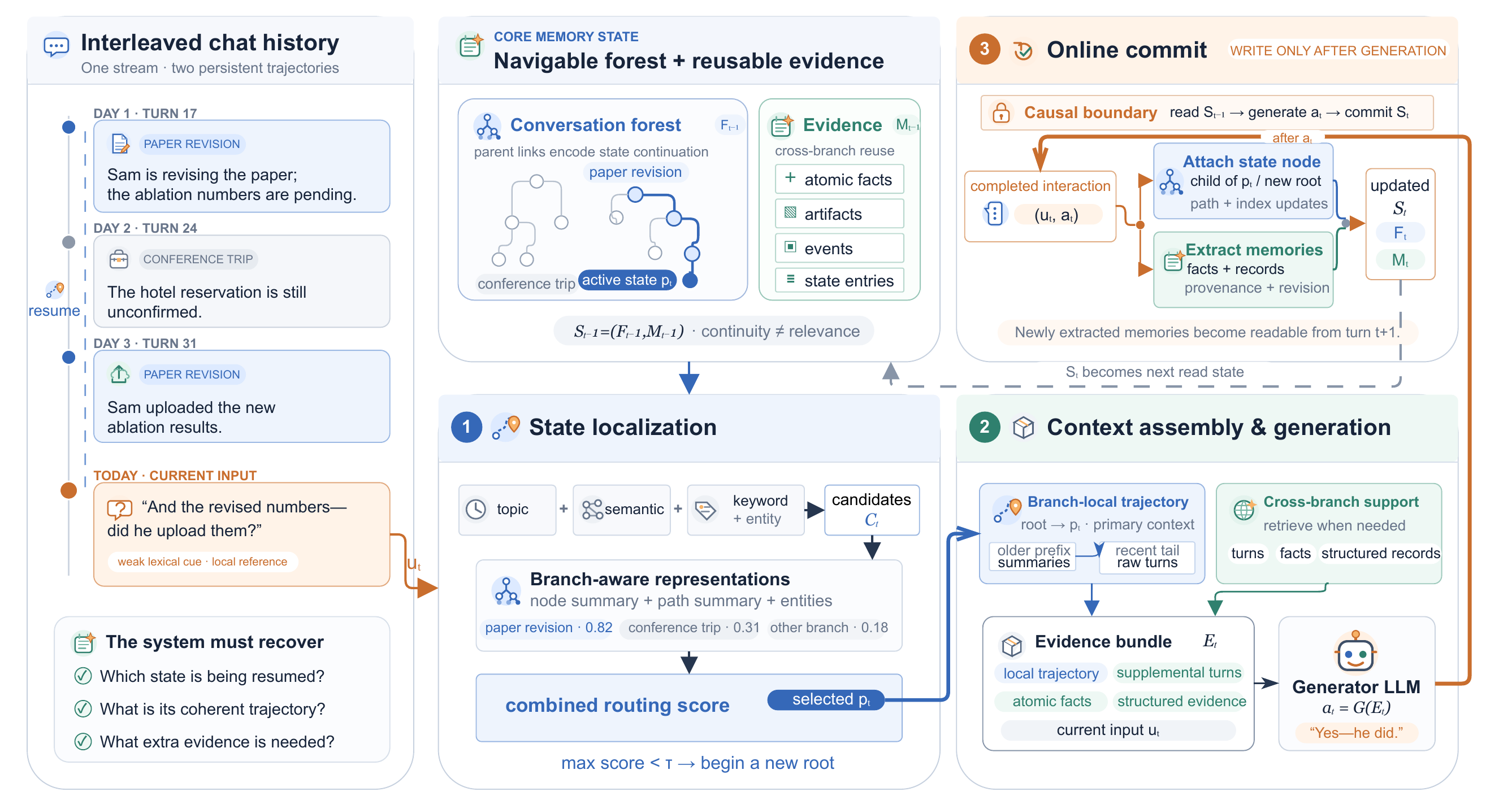}
\caption{Overview of \method{}. The committed memory state consists of
a conversation forest and a reusable evidence store. For each input,
\method{} localizes the resumed interaction state, reconstructs its
branch-local trajectory, and augments it with cross-branch evidence.
After generation, the completed interaction and extracted memories are
committed for use in subsequent turns.}
\label{fig:method}
\end{figure*}

\section{Navigable Memory Forest}
\label{sec:formulation}

We represent long-running conversational memory as a navigable forest of interaction states, preserving multiple trajectories that may be interrupted and later resumed. After turn $t$, the system maintains a committed memory state
\begin{equation}
S_t = (F_t, M_t),
\end{equation}
where $F_t$ is an interaction-state forest and $M_t$ is a reusable evidence store. Each node represents a committed local state, each parent edge represents state continuation, and each root-to-node path forms a locally coherent trajectory.

At turn $t$, $F_{t-1}$ determines the primary context in which the new input is interpreted, while $M_{t-1}$ supplies reusable facts and records across trajectory boundaries. The system reads only $S_{t-1}$; after generating the response, it attaches the completed interaction to the selected trajectory or begins a new root, and commits the updated state $S_t$. This separation preserves interaction continuity while allowing information reuse across branches.

\section{\method{}}
\label{sec:arbormem}

\method{} operationalizes the navigable memory forest through three stages, as illustrated in Figure~\ref{fig:method}. Given a new input $u_t$ and the committed state $S_{t-1}$, it first localizes the resumed interaction state, then assembles branch-local context and cross-branch evidence to generate $a_t$, and finally commits the completed interaction to produce $S_t$.

\subsection{Memory State}

\paragraph{Conversation forest.}
The conversation forest $\mathcal{F}_t$ represents interaction
continuity. Each node is created from one user--assistant interaction
and represents the local state established after that turn. It stores
the original interaction, node and path summaries, extracted entities
and keywords, a vector representation, and structural metadata including
its parent, root, depth, and global sequence position.

The forest grows online: each new interaction is attached to a
previously established state or inserted as a new root. A root-to-node
path forms a locally coherent trajectory that may later be resumed.
The topology therefore encodes discourse continuity rather than
semantic relevance alone; semantically related interactions need not
belong to the same local trajectory.

\paragraph{Reusable evidence store.}
The evidence store $\mathcal{M}_t$ contains information that may be
reused outside the trajectory in which it was introduced. It maintains
atomic facts and structured records. Atomic facts capture compact
information such as preferences, identifiers, dates, and project
details, while structured records represent assistant-generated
artifacts, events, and mutable state entries. Each item retains
provenance and revision metadata, enabling source tracing and the
filtering or deprioritization of obsolete information.

\subsection{State Localization}

Given a new input $u_t$ and the committed forest
$\mathcal{F}_{t-1}$, \method{} identifies the interaction state under which the input should be interpreted. Localization follows a priority cascade. Inputs with explicit continuation cues are attached directly to the active state, while explicit temporal-return cues are resolved through the global interaction timeline. These fast paths handle cases in which the intended state can be identified without general retrieval.

For other inputs, \method{} constructs a hybrid candidate set
$\mathcal{C}_t$ from three sources. Semantic recall retrieves candidate
nodes through FAISS vector search~\citep{johnson2017billionscalesimilaritysearchgpus}. Keyword recall queries lexical and
entity indexes using extracted content words, lemmatized forms, and
entity terms. When explicit topic anchors are detected, topic-matched
nodes are also added to the candidate set.

Each candidate node $v\in\mathcal{C}_t$ is then reranked against the
current input using a branch-aware representation. This representation
contains the candidate's topic key when available, its path summary,
node summary, and extracted entities. The cross-encoder therefore
evaluates not only the candidate node in isolation, but also the
interaction trajectory leading to it.

The implemented localization score is an additive combination of the
cross-encoder score and auxiliary routing signals:
\begin{equation} s_t(v) = s_{\mathrm{ce}}(v) +s_{\mathrm{key}}(v) +s_{\mathrm{rec}}(v) +s_{\mathrm{topic}}(v), \label{eq:localization-score} \end{equation}
where $s_t(v)$ denotes the localization score of candidate node $v$ for the current input $u_t$. Here, $s_{\mathrm{ce}}$ is the branch-aware cross-encoder reranking score, $s_{\mathrm{key}}$ is a rule-based keyword and entity adjustment, $s_{\mathrm{rec}}$ is a recency bias, and $s_{\mathrm{topic}}$ is an optional topic-anchor adjustment.

The parent state is selected as \begin{equation} p_t = \begin{cases} \displaystyle \operatorname*{arg\,max}_{v\in\mathcal{C}_t} s_t(v), & \text{if } \displaystyle \max_{v\in\mathcal{C}_t} s_t(v) > \tau, \\[6pt] \varnothing, & \text{otherwise}, \end{cases} \label{eq:parent-selection} \end{equation} where $\tau$ is the threshold for starting a new root. If the highest score exceeds $\tau$, the corresponding candidate is selected as the parent state, and its root-to-node path forms the branch-local context for the current input.

\subsection{Context Assembly and Generation}

If $p_t$ is defined, \method{} reconstructs the root-to-$p_t$ path as
the branch-local trajectory. If no candidate exceeds the localization
threshold, the trajectory is empty and the current interaction begins
a new root.

If the selected trajectory fits within the read budget, its original
turns are included directly. For longer trajectories, \method{}
compresses the older prefix into summaries while preserving a recent
tail of raw turns. This retains coherent local context without
introducing unrelated branches.

Because the branch-local trajectory may not contain all information
needed for generation, \method{} retrieves cross-branch support from
three sources: supplemental turns outside the active path, reusable
atomic facts, and structured records containing artifacts, events, and
state entries.

At inference time, \method{} assembles four memory channels together
with the current input:
\begin{equation}
E_t =
\bigl[
C_t^{\mathrm{local}};
E_t^{\mathrm{turn}};
E_t^{\mathrm{fact}};
E_t^{\mathrm{struct}};
u_t
\bigr],
\label{eq:evidence-bundle}
\end{equation}
where $C_t^{\mathrm{local}}$ is the branch-local trajectory and
$E_t^{\mathrm{turn}}$, $E_t^{\mathrm{fact}}$, and
$E_t^{\mathrm{struct}}$ contain supplemental turns, atomic facts, and
structured records, respectively. Cross-branch evidence is retrieved
through semantic and entity-based matching under the remaining read
budget.

Structured records are serialized into a dedicated evidence block that
preserves their types, content, provenance, and revision status,
separately from the branch-local context and atomic facts.

The response is then generated as
\begin{equation}
a_t = G(E_t).
\label{eq:generation}
\end{equation}
The branch-local trajectory supports state-consistent interpretation,
while cross-branch evidence supplies additional information required
for broader reasoning.

\subsection{Online Memory Commit}

After generation, \method{} incorporates the completed interaction into
the evolving memory state:
\begin{equation}
S_t =
\operatorname{Commit}
\bigl(S_{t-1},u_t,a_t\bigr).
\label{eq:online-commit}
\end{equation}

If $p_t$ is defined, a new state node is inserted as its child.
Otherwise, the interaction begins a new root. The system then updates
path summaries, semantic and keyword indexes, structural metadata, and
the active-state pointer.


Separate post-generation extractors identify reusable atomic facts and structured records from the completed interaction. 
Newly extracted items are linked to their source node, while revision metadata may mark earlier information as obsolete or superseded.

All extraction and memory updates occur after generation. Thus, $a_t$ depends only on the previously committed state $S_{t-1}$, and information extracted from $(u_t,a_t)$ becomes readable only from turn $t+1$. This preserves the causal boundary between memory reading and writing.

Complete implementation details, including routing hyperparameters, candidate construction, memory schemas, evidence allocation, and prompts, are provided in the supplementary material.

\section{BranchMemEval}
\label{sec:benchmark}

\paragraph{Scope and dimensions.}
Existing conversational-memory benchmarks evaluate factual retention, temporal and multi-hop reasoning, knowledge updates, and long histories, but do not explicitly isolate interference from interleaved and resumable interaction trajectories. We therefore introduce \benchmark{}, a controlled diagnostic benchmark for branch-structured
conversational memory. Most sessions fit within the 32K answer-time evidence budget, allowing the benchmark to focus on trajectory localization and state tracking rather than context overflow.
\benchmark{} evaluates five capabilities: B1 (\emph{backjump}) resumes an earlier trajectory after intervening dialogue; B2 (\emph{state update}) retrieves the latest value within the correct trajectory; B3 (\emph{assistant artifacts}) tests memory for assistant-generated lists and records; B4 (\emph{cross-branch bridging}) combines evidence across trajectories; and B5 (\emph{twin threads}) distinguishes structurally similar parallel agendas.

\paragraph{Construction.}
\benchmark{} contains 33 sessions and 100 in-dialogue questions across four session families jointly covering the five diagnostic dimensions, with three or four topical branches interleaved in each session. 
A programmatic planner specifies branches, facts, updates, artifacts, question positions, gold answers, and controlled
confounders, after which non-question turns are verbalized into natural
dialogue. Questions and canonical answers are derived from the
underlying structured metadata. Automatic validation checks that each
answer is supported by the preceding history, that neither the answer
nor its confounder is leaked in the question, and that parallel
trajectories remain distinguishable. Latent branch annotations are used
only for construction and analysis and are never exposed to evaluated
systems.

\paragraph{Online protocol.}
Sessions are replayed turn by turn, and each question must be answered using only the previously committed dialogue prefix. After an answer is produced, replay continues with the dataset's canonical turns rather than the generated response, preventing model outputs from altering the
subsequent history. 
Further construction, validation, and dataset details are provided in the supplementary material.
\begin{table*}[t]
\centering
\normalsize
\setlength{\tabcolsep}{3.0pt}
\renewcommand{\arraystretch}{1.25}
\begin{tabular}{lccccccccc}
\toprule
& \multicolumn{4}{c}{\textbf{Context and Retrieval}}
& \multicolumn{5}{c}{\textbf{Memory Systems}} \\
\cmidrule(lr){2-5}
\cmidrule(lr){6-10}
\textbf{Benchmark}
& \textbf{Full}
& \textbf{Recent}
& \textbf{BM25}
& \textbf{Summary}
& \textbf{Graphiti}
& \textbf{A-MEM}
& \textbf{Mem0}
& \textbf{LiCoMem.}
& \textbf{\method{}} \\
\midrule
LongMemEval
& 27.40
& 29.00
& 27.40
& 27.20
& 9.00
& 52.40
& \underline{59.40}
& 26.40
& \textbf{68.40} \\

LoCoMo
& 42.09
& 16.58
& \underline{42.84}
& 28.50
& 38.18
& 39.71
& 27.21
& 32.61
& \textbf{53.15} \\

BEAM 100K
& 22.27
& 18.64
& 37.05
& 43.00
& 44.38
& 44.98
& \underline{46.39}
& 36.74
& \textbf{49.75} \\

\benchmark{}
& 66.00
& 59.00
& 73.00
& 43.00
& 18.00
& 73.00
& \underline{76.00}
& 72.00
& \textbf{81.00} \\
\bottomrule
\end{tabular}
\caption{Answer accuracy (\%) on four conversational memory benchmarks. LoCoMo results cover Categories 1--4. The best result is shown in bold, and the strongest
baseline is underlined.}
\label{tab:main-results}
\end{table*}

\begin{table*}[t]
\centering
\normalsize
\setlength{\tabcolsep}{3.0pt}
\renewcommand{\arraystretch}{1.25}
\begin{tabular}{lccccccccc}
\toprule
& \multicolumn{4}{c}{\textbf{Context and Retrieval}}
& \multicolumn{5}{c}{\textbf{Memory Systems}} \\
\cmidrule(lr){2-5}
\cmidrule(lr){6-10}
\textbf{Budget}
& \textbf{Full}
& \textbf{Recent}
& \textbf{BM25}
& \textbf{Summary}
& \textbf{Graphiti}
& \textbf{A-MEM}
& \textbf{Mem0}
& \textbf{LiCoMem.}
& \textbf{\method{}} \\
\midrule
256
& 8.0
& 8.0
& 8.0
& 4.0
& 4.0
& 28.0
& \underline{38.0}
& 14.0
& \textbf{64.0} \\

512
& 8.0
& 8.0
& 8.0
& 6.0
& 8.0
& 34.0
& \underline{50.0}
& 12.0
& \textbf{74.0} \\

1K
& 8.0
& 8.0
& 8.0
& 12.0
& 18.0
& 50.0
& \underline{58.0}
& 10.0
& \textbf{74.0} \\

2K
& 8.0
& 8.0
& 8.0
& 30.0
& 36.0
& 56.0
& \textbf{72.0}
& 14.0
& \underline{68.0} \\

4K
& 8.0
& 8.0
& 10.0
& 50.0
& 52.0
& \underline{68.0}
& 60.0
& 14.0
& \textbf{70.0} \\

8K
& 10.0
& 8.0
& 12.0
& 56.0
& 56.0
& 62.0
& \underline{66.0}
& 16.0
& \textbf{76.0} \\

16K
& 16.0
& 16.0
& 22.0
& \underline{70.0}
& 68.0
& 60.0
& 66.0
& 22.0
& \textbf{72.0} \\

32K
& 28.0
& 26.0
& 42.0
& \underline{68.0}
& \textbf{76.0}
& 56.0
& 66.0
& 24.0
& \textbf{76.0} \\
\bottomrule
\end{tabular}
\caption{Answer accuracy (\%) on a fixed 50-question LongMemEval subset under different answer-time evidence budgets. Methods are grouped into context-and-retrieval baselines and memory systems. Bold indicates the best result at each budget, and underlining indicates the second-best result.}
\label{tab:read-budget}
\end{table*}

\section{Experiments}
\label{sec:experiments}

We evaluate \method{} as an online memory system for long-running conversations. 
Our experiments address five questions: 
(1) whether \method{} improves answer accuracy over long-context, retrieval-based, and existing agent-memory systems; 
(2) whether it selects effective evidence under constrained answer-time read budgets; 
(3) whether its online ingestion and query costs remain practical; 
(4) whether trajectory localization and branch-local context contribute to answer accuracy; 
and (5) how its evidence-processing components affect the final system.

\subsection{Experimental Setup}
\label{sec:setup}


\paragraph{Benchmarks.}
We evaluate \method{} on four conversational-memory benchmarks:
LongMemEval~\citep{wu2025longmemevalbenchmarkingchatassistants},
LoCoMo~\citep{maharana-etal-2024-evaluating},
BEAM 100K~\citep{tavakoli2026milliontokensbenchmarkingenhancing},
and \benchmark{}.
LongMemEval contains 500 questions covering information extraction,
multi-session reasoning, temporal reasoning, knowledge updates, and
abstention. For LoCoMo, we evaluate Categories 1--4, comprising 1,540
questions; Category 5 is excluded because it contains unanswerable
questions whose abstention handling is inconsistent across prior
baseline implementations. BEAM 100K contains 20 coherent conversations
and 400 questions across ten categories, with histories of approximately
100K tokens. \benchmark{} provides the controlled evaluation of
branch-structured conversational memory introduced above.

\paragraph{Baselines.}
We compare \method{} with eight baselines spanning direct context access, retrieval and compression, and structured agent memory.
Full-context and Recent provide chronological history under the context budget; BM25 retrieves relevant turns or chunks; and Session Summary uses model-generated session summaries. Structured memory baselines include Graphiti~\citep{rasmussen2025zeptemporalknowledgegraph}, A-MEM~\citep{NEURIPS2025_19909c36}, Mem0~\citep{chhikara2025mem0buildingproductionreadyai}, and LiCoMemory~\citep{huang-etal-2026-licomemory}.

When supported, all methods use the same answer model and answer-time evidence budget. A-MEM uses its accelerated ingestion path, Mem0 uses a retrieval-oriented configuration, and LiCoMemory's retrieved evidence is passed to the shared answer model. 



\paragraph{Implementation and protocol.}
The main experiments use Qwen3-30B-A3B-Instruct-2507 for answer
generation and model-based memory operations
\citep{yang2025qwen3technicalreport}. We use
BGE-M3~\citep{chen-etal-2024-m3} for embedding and
BGE-Reranker-v2-M3 for reranking. When supported, all methods share the
same answer model, decoding configuration, and approximately 32K
answer-time evidence budget.

For LongMemEval, LoCoMo, and BEAM 100K, conversations are ingested
chronologically and questions are presented after the complete history.
Each case maintains an independent memory state. For \benchmark{},
questions occur within the dialogue and are answered using only the
previously committed prefix, after which the canonical dialogue
continues.

\paragraph{Evaluation.}
We report answer accuracy using each benchmark's corresponding evaluation protocol. LongMemEval predictions are evaluated with a local correctness judge against the reference answers, while LoCoMo and BEAM 100K use their benchmark-specific scoring procedures. \benchmark{} uses rule-based normalized accuracy with token-boundary matching.
Because the main experiment and subsequent analyses differ in data subsets, model sizes, evidence budgets, and memory configurations, results are directly comparable only within the same table.

Detailed baseline configurations and reproducibility information are provided in the supplementary material.

\subsection{Main Results}
\label{sec:main-results}

Table~\ref{tab:main-results} reports end-to-end answer accuracy.
\method{} achieves the best result on all four benchmarks, outperforming the strongest baseline by 9.00 points on LongMemEval, 10.31 points on LoCoMo, 3.36 points on BEAM 100K, and 5.00 points on \benchmark{}.

The largest improvements occur on LongMemEval and LoCoMo, where questions often require information distributed across sessions or conversational contexts. On \benchmark{}, full-context remains competitive at 66.00\%, yet \method{} outperforms it by 15.00 points and exceeds the strongest memory baseline by 5.00 points. This result shows that exposing the available history alone is insufficient for reliable reasoning over interleaved interaction threads.

Full-context and recent-window methods depend on the amount and position of visible history, while retrieval and summary methods may recover relevant information without preserving its local conversational context. By combining branch-local context with reusable evidence, \method{} achieves consistent gains across natural multi-session, very-long-context, and controlled branch-structured evaluations.

\subsection{Read-Budget Analysis}
\label{sec:read-budget}

We next examine how effectively each method selects evidence when the answer model is given a limited read budget. Due to the computational cost of evaluating all methods across multiple budgets, we use a fixed subset of 50 LongMemEval questions. We evaluate eight answer-time evidence budgets: 256, 512, 1K, 2K, 4K, 8K, 16K, and 32K tokens. All runs use Qwen3-30B-A3B-Instruct-2507.


For each answer-time evidence budget, all methods are limited to the same amount of evidence supplied to the answer model. 
\method{} ingests the complete interaction history and constrains the combined branch-local context and retrieved evidence to the specified budget. 
Other memory systems similarly construct their memory representations before selecting evidence under the same answer-time limit. 
This experiment therefore evaluates how effectively each method constructs a compact generation context under constrained read budgets. 


\method{} achieves the best result at six of the eight budgets and ties for the best result at 32K. Its advantage is particularly pronounced under highly constrained budgets: from 256 to 1K tokens, it exceeds the strongest competing method by 16--26 percentage points. As the budget increases, summary- and graph-based methods generally benefit from access to more evidence, while \method{} remains strong across the full range of evidence budgets. These results indicate that \method{} can construct effective generation contexts under both highly constrained and relatively large answer-time budgets.

\subsection{Efficiency and Latency}
\label{sec:efficiency}

We compare the efficiency of \method{}, Mem0, and A-MEM on the same fixed 50-question LongMemEval subset. The histories contain 12,394 turns in total, averaging 247.88 turns per case. All methods use Qwen3-4B-Instruct-2507 through the same local inference service and are executed sequentially under identical 4K memory-operation and 2K answer-time evidence budgets.

\begin{table*}[t]
\centering
\normalsize
\setlength{\tabcolsep}{5.0pt}
\renewcommand{\arraystretch}{1.25}
\begin{tabular}{lrrrrrr}
\toprule
& \multicolumn{1}{c}{\textbf{Memory Construction}}
& \multicolumn{3}{c}{\textbf{Query Time}}
& \multicolumn{2}{c}{\textbf{End-to-End}} \\
\cmidrule(lr){2-2}
\cmidrule(lr){3-5}
\cmidrule(lr){6-7}
\textbf{Method}
& \textbf{Ingest/Turn $\downarrow$}
& \textbf{Query Prep. $\downarrow$}
& \textbf{TTFT $\downarrow$}
& \textbf{Full Query $\downarrow$}
& \textbf{Total Time $\downarrow$}
& \textbf{Case Throughput $\uparrow$} \\
& \textbf{(s)}
& \textbf{(s)}
& \textbf{(s)}
& \textbf{(s)}
& \textbf{(min)}
& \textbf{(q/h)} \\
\midrule
\method{}
& \textbf{1.363}
& 0.213
& 0.381
& 0.451
& \textbf{287.0}
& \textbf{10.45} \\

Mem0
& 1.543
& \textbf{0.033}
& \textbf{0.153}
& 0.268
& 319.4
& 9.39 \\

A-MEM
& 3.091
& 0.045
& 0.198
& \textbf{0.236}
& 639.3
& 4.69 \\
\bottomrule
\end{tabular}
\caption{Matched efficiency comparison on a fixed 50-question
LongMemEval subset. All methods use the same local inference service
and sequential processing. Lower is better for latency and runtime,
while higher is better for throughput.}
\label{tab:efficiency}
\end{table*}

\method{} has the lowest memory-construction cost, requiring 1.363
seconds per ingested turn, compared with 1.543 seconds for Mem0 and
3.091 seconds for A-MEM. It is therefore moderately faster than Mem0
and more than twice as fast as A-MEM during ingestion, which is
important when long-running conversations require hundreds of turns to
be incorporated into memory.

Mem0 and A-MEM have lower query latency, while \method{} additionally
performs state localization and branch reconstruction before generation.
Nevertheless, its full-query latency remains below half a second.
Because ingestion dominates the total runtime, \method{} completes the
evaluation fastest and achieves the highest end-to-end case throughput
at 10.45 questions per hour, compared with 9.39 for Mem0 and 4.69 for
A-MEM.

\subsection{Component Ablations}
\label{sec:ablation}

We evaluate state localization and four evidence-processing components on the same fixed 50-question LongMemEval subset used in the read-budget analysis. All configurations ingest the complete history and use a 32K answer-time evidence budget. 
The reference configuration includes state localization and the resulting branch-local trajectory, keyword and entity routing, atomic-fact retrieval, structured-record extraction, and a dedicated structured evidence block.

The state-localization ablation disables branch routing and node selection, and therefore does not construct branch-local context, while retaining globally retrieved atomic facts and structured records. Each remaining ablation removes one component while leaving the others unchanged. We evaluate both Qwen3-30B-A3B-Instruct-2507 and Qwen3-4B-Instruct-2507. Because the two settings use different automatic judges, results should be compared within each model column.

\begin{table}[t]
\centering
\small
\renewcommand{\arraystretch}{1.25}
\setlength{\tabcolsep}{2.5pt}
\begin{tabular*}{\columnwidth}{
    @{\extracolsep{\fill}}
    lcc
    @{}
}
\toprule
\textbf{Configuration}
&
\shortstack[c]{
    \footnotesize\bfseries Qwen3-30B-A3B-\\[-0.5pt]
    \footnotesize\bfseries Instruct-2507
}
&
\shortstack[c]{
    \footnotesize\bfseries Qwen3-4B-\\[-0.5pt]
    \footnotesize\bfseries Instruct-2507
} \\
\midrule

Reference configuration
& \textbf{82.0}
& \textbf{48.0} \\

\quad w/o state localization
& 70.0 {\scriptsize ($-14.6\%$)}
& 46.0 {\scriptsize ($-4.2\%$)} \\

\quad w/o keyword/entity routing
& 74.0 {\scriptsize ($-9.8\%$)}
& 44.0 {\scriptsize ($-8.3\%$)} \\

\quad w/o atomic-fact retrieval
& 62.0 {\scriptsize ($-24.4\%$)}
& 42.0 {\scriptsize ($-12.5\%$)} \\

\quad w/o structured extraction
& 68.0 {\scriptsize ($-17.1\%$)}
& 40.0 {\scriptsize ($-16.7\%$)} \\

\quad w/o structured evidence
& 68.0 {\scriptsize ($-17.1\%$)}
& 44.0 {\scriptsize ($-8.3\%$)} \\

\bottomrule
\end{tabular*}
\caption{ Accuracy (\%) on a fixed 50-question LongMemEval subset. The state-localization ablation disables branch routing and branch-local context construction while retaining globally retrieved atomic facts and structured records. Parenthetical values indicate relative drops from the reference configuration.}
\label{tab:ablation}
\end{table}

Removing state localization reduces accuracy from 82.0\% to 70.0\% with the 30B model and from 48.0\% to 46.0\% with the 4B model. The 12-point decrease in the 30B setting shows that globally retrieved facts and structured records cannot fully replace the coherent context obtained by locating and restoring the resumed interaction state.

All evidence-processing ablations also reduce accuracy. Atomic-fact retrieval has the largest observed effect in the 30B setting, where its removal lowers accuracy to 62.0\%, while structured-record extraction has the largest effect in the 4B setting, lowering accuracy to 40.0\%. The remaining ablations likewise produce consistent decreases. Overall, the results indicate complementary contributions from state localization and branch-local context, keyword- and entity-based routing, reusable facts, structured-record extraction, and structured evidence presentation.




\section{Discussion}
\label{sec:discussion}

\paragraph{Conversational state and forest structure.}
\method{} separates conversational continuation from semantic
relevance: the localized trajectory provides the primary context for
interpreting a turn, while reusable evidence can be retrieved across
branches. Its single-parent forest yields a clear primary context and
supports compact path-based context construction. However, ambiguous or
multi-intent turns may correspond to multiple states, and routing errors
may propagate into subsequent memory updates. Future extensions could
model localization uncertainty or introduce secondary links while
preserving a navigable primary trajectory.

\paragraph{Limitations of current memory systems.}
Like other structured memory systems, \method{} depends on reliable
extraction, summarization, retrieval, updating, and revision. Incorrect,
stale, or conflicting memories may persist, while richer memory
construction introduces additional write-time cost. Existing benchmarks
also focus mainly on answering questions over fixed histories and do not
fully capture open-ended interaction, user-requested deletion, or the
cumulative effects of incorrect memory updates. Future evaluation should
therefore consider localization reliability, revision quality,
uncertainty, and long-term error propagation in addition to answer
accuracy.
\section{Conclusion}
\label{sec:conclusion}

We presented \method{}, an online framework organizing long-running conversations as navigable forests of interaction states. By localizing each input to a coherent trajectory and augmenting it with reusable cross-branch evidence, \method{} supports state-consistent interpretation and long-term information reuse. 
We also introduced \benchmark{}, a controlled benchmark for interleaved, resumable trajectories.
Experiments on LongMemEval, LoCoMo, BEAM 100K, and \benchmark{} demonstrate consistent improvements over long-context, retrieval-based, and structured memory baselines, particularly under constrained read budgets. These results establish interaction-state localization as a basis for reliable long-term memory and provide a controlled setting for studying branch-structured challenges.
\newpage
\bibliography{references}

\clearpage

\appendix

\section{ArborMem Implementation Details}
\label{app:implementation}

This section reports the implementation choices needed to reproduce
\method{}. We omit the method definitions already introduced in the
main paper and focus on the online execution order, context packing,
and reference hyperparameters.

\subsection{Online Execution}

At turn $t$, \method{} retrieves evidence only from the previously
committed state $S_{t-1}$. It first resolves explicit continuation or
temporal-return cues and otherwise retrieves and reranks candidate
interaction states. The selected root-to-parent trajectory is combined
with reusable factual and structured evidence to answer the current
input. After generating $a_t$, the completed interaction is attached
below the selected parent or inserted as a new root, and the associated
summaries, indexes, and reusable records are updated. Consequently,
information extracted from $(u_t,a_t)$ becomes readable only from turn
$t+1$.

\subsection{Retrieval and Context Settings}

Table~\ref{tab:app-arbormem-settings} summarizes the reference implementation settings used in \method{}. 

\begin{table*}[t]
\centering
\normalsize
\setlength{\tabcolsep}{6pt}
\renewcommand{\arraystretch}{1.22}
\begin{tabularx}{\textwidth}{
    p{0.22\textwidth}
    p{0.23\textwidth}
    X
}
\toprule
\textbf{Component} & \textbf{Setting} & \textbf{Reference configuration} \\
\midrule

State recall
& Semantic candidates
& Up to 20 interaction-state nodes. \\

& Topic candidates
& Up to 12 additional candidates per detected topic anchor. \\

\midrule

State localization
& New-root threshold
& The best candidate must strictly exceed $\tau = 8.0$. \\

& Keyword/entity adjustment
& $1.0$ for the first informative overlap and $0.5$ for each additional
overlap; terms appearing in more than 40\% of nodes are ignored. \\

& Recency adjustment
& Bounded auxiliary bonus with a maximum value of $1.0$. \\

& Topic adjustment
& $8.0 \times$ topic-match score for matches above $0.5$, and $-6.0$
otherwise. \\

\midrule

Branch context
& Recent raw tail
& Four interaction-state nodes. \\

& Maximum branch budget
& 16,384 tokens. \\

\midrule

Reusable evidence
& Atomic-fact retrieval
& Up to 10 facts with a minimum semantic similarity of $0.4$. \\

& Fact key expansion
& The fact and its source interaction are jointly represented for
retrieval; only the compact fact is passed to the answer model. \\

& Supplemental raw turns
& Supported by the architecture but assigned zero budget in the
reported experiments. \\

\bottomrule
\end{tabularx}
\caption{Reference implementation settings of \method{}.}
\label{tab:app-arbormem-settings}
\end{table*}

When the selected trajectory fits within the branch budget, its
interactions are retained in full. Otherwise, the four most recent
nodes remain as raw user--assistant turns, while earlier nodes are
represented by their summaries. Atomic facts and structured records
are retrieved independently and appended without replacing the selected
trajectory as the primary conversational context.

\subsection{Memory Commit}

After generation, separate extractors update the node summary, path
summary, entities, atomic facts, and structured records. All extracted
records retain provenance to their source interaction. When a newly
extracted fact or mutable state refers to an existing attribute, a
revision check determines whether the records can coexist or whether
the earlier value should be treated as superseded. Superseded records
remain traceable but may be filtered or deprioritized during subsequent
retrieval.

\section{BranchMemEval Construction and Validation}
\label{app:bme}

BranchMemEval is a controlled diagnostic benchmark for evaluating
memory under interleaved, resumable, and structurally similar
interaction trajectories. Rather than increasing transcript length
alone, it tests whether a system can identify the relevant interaction
state and retrieve the correct information associated with that state.
This section reports the benchmark composition, construction controls,
validation procedure, and evaluation protocol.

\subsection{Benchmark Composition}
\label{app:bme-composition}

The formal BranchMemEval set contains 33 sessions, 100 in-dialogue
questions, and 1,442 non-question interactions. Each session contains
three or four interleaved topical branches and 42--52 total turns, with
an average of 46.73 turns. Table~\ref{tab:bme-composition} summarizes
the diagnostic dimensions and construction families.

\medskip
\noindent
\begin{minipage}{\columnwidth}
\centering
\small
\setlength{\tabcolsep}{4pt}
\renewcommand{\arraystretch}{1.15}

\textbf{(a) Diagnostic dimensions}

\vspace{3pt}

\begin{tabularx}{\columnwidth}{
    @{}
    l
    >{\raggedright\arraybackslash}X
    r
    @{}
}
\toprule
\textbf{Dim.}
&
\textbf{Controlled capability}
&
\textbf{Qs.}
\\
\midrule

B1
&
Resume an earlier trajectory after intervening dialogue.
&
26
\\

B2
&
Retrieve the latest value from the correct trajectory.
&
16
\\

B3
&
Recall a referenced assistant-generated artifact.
&
24
\\

B4
&
Combine evidence introduced in two trajectories.
&
14
\\

B5
&
Disambiguate parallel same-schema trajectories.
&
20
\\

\midrule
\multicolumn{2}{r}{\textbf{Total}}
&
\textbf{100}
\\
\bottomrule
\end{tabularx}

\vspace{8pt}

\textbf{(b) Construction families}

\vspace{3pt}

\begin{tabular*}{\columnwidth}{
    @{\extracolsep{\fill}}
    lccc
    @{}
}
\toprule
\textbf{Family}
&
\textbf{Target}
&
\textbf{Sess.}
&
\textbf{Qs.}
\\
\midrule

Backjump & B1, B2   & 8  & 32 \\
Artifact & B3       & 8  & 24 \\
Twin     & B5 (+B1) & 10 & 30 \\
Bridge   & B4       & 7  & 14 \\

\midrule
\textbf{Total}
&
--
&
\textbf{33}
&
\textbf{100}
\\
\bottomrule
\end{tabular*}

\captionof{table}{Composition of BranchMemEval. Panel (a) reports the
distribution of diagnostic questions, while Panel (b) reports the
session families used during construction.}
\label{tab:bme-composition}
\end{minipage}
\medskip

The construction families and diagnostic dimensions are not strictly
one-to-one. The backjump family jointly covers trajectory resumption
and state updates, while some twin-thread sessions also contain
backjump questions. The artifact, twin, and bridge families primarily
target B3, B5, and B4, respectively.

Branch identities, slot annotations, and diagnostic labels are used
only for benchmark construction, validation, and aggregate analysis.
They are never provided to evaluated memory systems.

\subsection{Construction and Validation}
\label{app:bme-construction}

BranchMemEval separates structural planning from natural-language
realization. A programmatic planner first specifies the branch
structure, factual slots, state updates, assistant-generated artifacts,
controlled distractors, question positions, gold answers, and
confounding answers. A language model then verbalizes the non-question
turns into natural user--assistant interactions. The formal set uses
DeepSeek-V4-Pro for this realization stage.

Question turns and canonical responses remain grounded in the
programmatic specification rather than being independently generated
by the realization model. Table~\ref{tab:bme-validation} summarizes the
principal construction and validation controls.

\medskip
\noindent
\begin{minipage}{\columnwidth}
\centering
\small
\setlength{\tabcolsep}{4pt}
\renewcommand{\arraystretch}{1.15}

\begin{tabularx}{\columnwidth}{
    @{}
    p{0.30\columnwidth}
    >{\raggedright\arraybackslash}X
    @{}
}
\toprule
\textbf{Stage}
&
\textbf{Control}
\\
\midrule

Structural planning
&
Defines branch assignments, factual slots, updates, artifacts,
distractors, question positions, gold answers, and confounders.
\\

Dialogue realization
&
Verbalizes only the non-question turns while preserving the planned
interaction structure.
\\

Answer grounding
&
Derives questions and canonical responses from the structured
metadata rather than from free-form generation.
\\

Evidence validity
&
Requires each gold answer to be supported by an interaction preceding
the corresponding question.
\\

Leakage prevention
&
Ensures that neither the gold answer nor the confounder appears in the
question and removes construction metadata that could reveal the
answer.
\\

Dimension checks
&
Verifies valid old--new pairs for B2, hidden linking values for B4,
and explicit instance anchors for B5.
\\

\bottomrule
\end{tabularx}

\captionof{table}{Construction and automatic validation controls used
in BranchMemEval.}
\label{tab:bme-validation}
\end{minipage}
\medskip

Each question is paired with a structurally plausible confounder. For
example, a question about one person's travel budget may use the budget
from a parallel person's agenda as the confounder. This design
distinguishes failure to retrieve the requested value from confusion
between similar trajectories.

A deterministic normalization stage is applied only to question turns
and canonical assistant responses. The 1,442 non-question interactions
retain their language-model-realized dialogue text.

\subsection{Online Evaluation and Scoring}
\label{app:bme-evaluation}

Each session is replayed chronologically. Ordinary user turns and their
canonical assistant responses are committed to the evaluated memory
system. When a question turn is reached, the current dialogue prefix is
frozen and the system produces an answer using only previously
committed information.

After scoring, replay continues with the dataset's canonical assistant
response rather than the evaluated model's prediction. This prevents an
incorrect prediction from modifying the history available to later
questions and ensures that all systems observe the same subsequent
trajectory.

BranchMemEval uses normalized rule-based accuracy rather than an LLM
judge. Predictions are case-folded; punctuation and English articles
are removed; whitespace and numerical commas are normalized; and
\texttt{\&} is treated as equivalent to \texttt{and}. Matching is
performed at token boundaries to avoid accidental substring matches.

Let $g$ denote the normalized gold answer and $c$ the normalized
confounder. Table~\ref{tab:bme-scoring} defines the four possible
outcomes.

\medskip
\noindent
\begin{minipage}{\columnwidth}
\centering
\small
\setlength{\tabcolsep}{5pt}
\renewcommand{\arraystretch}{1.15}

\begin{tabularx}{\columnwidth}{
    @{}
    >{\raggedright\arraybackslash}X
    p{0.35\columnwidth}
    @{}
}
\toprule
\textbf{Matched content}
&
\textbf{Outcome}
\\
\midrule

Contains $g$ but not $c$
&
Correct
\\

Contains $c$ but not $g$
&
Trajectory confusion
\\

Contains both $g$ and $c$
&
Ambiguous
\\

Contains neither $g$ nor $c$
&
Miss
\\

\bottomrule
\end{tabularx}

\captionof{table}{Rule-based scoring outcomes in BranchMemEval.}
\label{tab:bme-scoring}
\end{minipage}
\medskip

The primary metric is overall answer accuracy. Per-dimension accuracy
and the three error categories are retained for diagnostic analysis.

\section{Baseline and Experimental Configurations}
\label{app:experiments}

This section provides the configurations required to reproduce the
reported results. We omit the high-level descriptions already given in
the main paper and focus on shared controls, benchmark-specific
protocols, baseline implementation choices, and the settings used in
the supplementary analyses.

\subsection{Shared Settings and Evaluation Protocols}
\label{app:shared-settings}

The shared model and evidence settings used in the main experiments are
summarized in Table~\ref{tab:app-shared-settings}.

\medskip
\noindent
\begin{minipage}{\columnwidth}
\centering
\small
\setlength{\tabcolsep}{4pt}
\renewcommand{\arraystretch}{1.13}

\begin{tabularx}{\columnwidth}{
    @{}
    p{0.38\columnwidth}
    >{\raggedright\arraybackslash}X
    @{}
}
\toprule
\textbf{Setting} & \textbf{Configuration} \\
\midrule

Answer and memory model
&
Qwen3-30B-A3B-Instruct-2507.
\\

Embedding model
&
BGE-M3.
\\

Reranker
&
BGE-Reranker-v2-M3.
\\

Maximum model context
&
40,960 tokens.
\\

Answer-time evidence
&
At most 30,968 tokens.
\\

Maximum answer generation
&
512 tokens.
\\

Retrieval candidates
&
Up to 50 memory units before final evidence packing.
\\

Intermediate prefilter
&
At most 61,936 tokens when required.
\\

\bottomrule
\end{tabularx}

\captionof{table}{Shared settings used in the main experiments.}
\label{tab:app-shared-settings}
\end{minipage}
\medskip

The answer-time evidence limit applies to the complete evidence supplied
to the generator. The smaller branch-context limit reported in
Table~\ref{tab:app-arbormem-settings} applies only to the localized
trajectory before atomic facts and structured evidence are added.

Whenever supported, all methods use the same answer model, decoding
configuration, evidence limit, and answer template. Evidence retrieved
by an external memory system is passed to the shared answer model rather
than evaluated with a method-specific generator.

\paragraph{LongMemEval.}
All 500 questions are evaluated independently. Each question uses an
isolated memory state, and the memory is reconstructed independently
even when multiple questions share the same underlying conversation.
Predictions are compared with the reference answer using a local
correctness judge following the benchmark evaluation format.

\paragraph{LoCoMo.}
Memory is constructed once for each conversation and reused across its
associated questions. We evaluate Categories 1--4, comprising 1,540
questions, using the category-specific scoring procedures. Category 5
is excluded because it contains unanswerable questions whose abstention
handling is not consistent across the available baseline
implementations.

\paragraph{BEAM 100K.}
Memory is constructed once for each of the 20 conversations and reused
for its associated questions. All 400 questions are evaluated using the
benchmark-specific per-item evaluation procedure.

\paragraph{BranchMemEval.}
Each session is replayed chronologically. Questions are answered using
only the previously committed conversation prefix, after which replay
continues with the canonical assistant response. The complete online
evaluation and rule-based scoring protocol is described in
Section~\ref{app:bme-evaluation}.

\subsection{Baseline Configurations}
\label{app:baseline-details}

Each independent benchmark case uses a separate memory instance. Unless
otherwise specified, all baselines expose their retrieved evidence to
the shared answer model under the common answer-time budget.

\paragraph{Full-context.}
This baseline uses raw chronological history without constructing an
external memory. When the complete history exceeds the available
budget, the newest turns are retained and then presented to the answer
model in chronological order.

\paragraph{Recent.}
Recent retains only the latest portion of the conversation, independent
of the current question. For BranchMemEval, it considers up to the most
recent 32 turns before packing them under the shared evidence budget.

\paragraph{BM25.}
BM25 indexes raw turns or benchmark-defined conversation chunks and
retrieves the highest-scoring units for the current question. Up to 50
retrieved units are considered before final packing. The BEAM adapter
uses the benchmark turn-chunk representation and retrieves up to 32
chunks.

\paragraph{Session Summary.}
This baseline maintains model-generated summaries of conversation
sessions. For datasets with explicit session boundaries, one summary is
maintained for each session. During BranchMemEval replay, the summary is
updated every eight interactions.

\paragraph{Graphiti.}
Graphiti stores entities, relations, events, and temporal associations
in its graph representation. Retrieved graph evidence and its
associated text are serialized and passed to the shared answer model.
During BranchMemEval replay, graph updates are performed every eight
interactions.

\paragraph{A-MEM.}
A-MEM is evaluated through a local adapter that preserves its
memory-note and association-based representation. The accelerated
ingestion path is used for full-benchmark evaluation, and retrieved
memory notes are passed to the shared answer model.

\paragraph{Mem0.}
Mem0 uses an independent retrieval-oriented memory store for each
evaluation case, with BGE-M3 for semantic retrieval. It does not use
\method{}'s keyword, topic, or branch-routing indexes.

\paragraph{LiCoMemory.}
Conversation histories are divided into approximately 8K-token chunks
and processed using LiCoMemory's native memory-graph construction. Its
native answer generator is disabled so that retrieved evidence can be
evaluated with the shared answer model. The answer-time limit applies
to the evidence returned for the current question rather than to the
complete constructed graph.

\paragraph{\method{}.}
Interactions are ingested sequentially using the reference
configuration reported in Section~\ref{app:implementation}. The full
configuration enables state localization, branch-local context,
keyword and entity routing, fact key expansion, atomic-fact retrieval,
and structured artifact, event, and mutable-state evidence.
Supplemental cross-branch raw turns are assigned zero budget in the
reported experiments.

For all retrieval and external-memory methods, the final evidence
passed to the generator is constrained by the shared answer-time limit.

\subsection{Supplementary Analysis Protocols}
\label{app:analysis-protocols}

The read-budget, efficiency, and component analyses use the settings in
Table~\ref{tab:app-analysis-settings}. Results should be compared only
within their corresponding experiments because the analyses use
different model sizes and evidence limits.

\medskip
\noindent
\begin{minipage}{\columnwidth}
\centering
\small
\setlength{\tabcolsep}{4pt}
\renewcommand{\arraystretch}{1.14}

\begin{tabularx}{\columnwidth}{
    @{}
    p{0.27\columnwidth}
    >{\raggedright\arraybackslash}X
    @{}
}
\toprule
\textbf{Analysis} & \textbf{Configuration} \\
\midrule

Read budget
&
A fixed 50-question LongMemEval subset sampled with seed 42;
Qwen3-30B-A3B-Instruct-2507; budgets of 256, 512, 1K, 2K, 4K, 8K,
16K, and 32K tokens. The 32K condition uses the exact 30,968-token
evidence limit.
\\

Efficiency
&
The same 50-question subset, containing 12,394 ingested turns
(247.88 per case); Qwen3-4B-Instruct-2507; sequential execution through
the same local vLLM service; 4K memory-operation and 2K answer-time
evidence budgets.
\\

Component ablation
&
The same 50-question subset; Qwen3-30B-A3B-Instruct-2507 and
Qwen3-4B-Instruct-2507; complete history ingestion and a 32K
answer-time evidence budget. Supplemental raw turns remain disabled.
\\

\bottomrule
\end{tabularx}

\captionof{table}{Configurations used in the supplementary analyses.}
\label{tab:app-analysis-settings}
\end{minipage}
\medskip

In the read-budget analysis, every method first ingests the complete
history. The experimental variable is the amount of evidence exposed
to the answer model. External memory methods may use an intermediate
prefilter of up to twice the target budget, but the final generator
input is limited to the same answer-time budget.

The efficiency comparison reports ingestion time per interaction,
query-preparation time, time to first token, complete query latency,
total runtime, and completed questions per hour. Query preparation
includes memory retrieval, state localization when applicable, and
evidence serialization.

The component analysis evaluates the following variants:

\begin{itemize}
    \item \textbf{w/o state localization}: removes parent selection and
    branch-local context while retaining globally retrieved atomic facts
    and structured records;
    \item \textbf{w/o keyword/entity routing}: removes the lexical
    routing channel;
    \item \textbf{w/o atomic-fact retrieval}: removes retrieved facts
    from the evidence bundle;
    \item \textbf{w/o structured extraction}: disables artifact, event,
    and mutable-state extraction;
    \item \textbf{w/o structured evidence}: retains the extracted
    records but removes their dedicated generation-time evidence block.
\end{itemize}

Each variant changes only the specified component. Because the two model
settings use different automatic judges, ablation results should be
compared within each model column rather than across model sizes.

\subsection{Isolation and Compute Environment}
\label{app:experimental-isolation}

Memory is never shared across independent benchmark cases. All
conversation turns are ingested chronologically, and no future turn is
available when an in-dialogue question is answered. Answer formatting
and normalization are held fixed within each benchmark.

The main experiments are conducted on a server equipped with eight
NVIDIA A800 accelerators, each with 80\,GB of memory. Seven A800s host
independent model-serving endpoints, while the remaining A800 hosts the
embedding and reranking models. The matched efficiency comparison uses
a separate sequential configuration with a single tensor-parallel
vLLM service. Its latency measurements are therefore reported
separately from those of the parallel main evaluation.

\lstdefinestyle{arbormemprompt}{
  basicstyle=\ttfamily\small,
  breaklines=true,
  breakatwhitespace=false,
  columns=fullflexible,
  keepspaces=true,
  showstringspaces=false,
  literate={✓}{{$\checkmark$}}1
           {—}{{--}}1
           {→}{{$\rightarrow$}}1
           {…}{{\ldots}}1
}
\newtcblisting{arbormempromptbox}[1]{
  enhanced,
  breakable,
  listing only,
  listing engine=listings,
  listing options={style=arbormemprompt},
  title={#1},
  fonttitle=\bfseries,
  colback=white,
  colframe=black!45,
  colbacktitle=black!5,
  coltitle=black,
  boxrule=0.5pt,
  arc=1mm,
  left=1.5mm,
  right=1.5mm,
  top=1mm,
  bottom=1mm,
  before skip=8pt,
  after skip=8pt
}

\clearpage
\onecolumn

\section{Prompts and Memory Schemas}
\label{app:prompts}

This section reports the prompts used by \method{} in the reported
experiments. Static prompts are reproduced verbatim from the
implementation. For dynamically assembled instructions, we report all
possible instruction lines and mark runtime-inserted values using angle
brackets. Only visual line wrapping and placeholder formatting are
changed.

We include prompts used for answer generation, retrieval reranking,
post-generation memory commit, and BranchMemEval-specific query
processing. Benchmark scoring directives are omitted because they
belong to the evaluation harness rather than the memory system.

\subsection{Prompt Usage and Runtime Assembly}
\label{app:prompt-usage}

Table~\ref{tab:prompt-usage} summarizes the role and invocation
condition of each included prompt. The artifact, event, and mutable-state
extractors are enabled in the full configuration used for the main
LongMemEval, LoCoMo, and BEAM experiments. The attribute-tag predictor
is used only by the BranchMemEval evaluation path.

\begin{table}[t]
\centering
\small
\setlength{\tabcolsep}{5pt}
\renewcommand{\arraystretch}{1.15}
\begin{tabularx}{\textwidth}{
    @{}
    p{0.18\textwidth}
    p{0.27\textwidth}
    >{\raggedright\arraybackslash}X
    @{}
}
\toprule
\textbf{Stage}
&
\textbf{Prompt or Instruction}
&
\textbf{Purpose and Invocation}
\\
\midrule

Answer generation
&
\texttt{system\_anchor}
&
Included as the first system message for every answer request. It
defines the evidence hierarchy, temporal behavior, conflict resolution,
and response style.
\\

&
\texttt{chain\_of\_note}
&
Used when atomic facts are retrieved. It instructs the model to inspect
facts, update evidence, branch-local history, and structured evidence.
\\

&
\texttt{chain\_of\_note\_no\_facts}
&
Used when no atomic facts are retrieved but conversational or
structured evidence remains available.
\\

&
Dynamic answer-format instructions
&
Adds only the rules corresponding to the detected query type, such as
current-value, ordinal, counting, or cross-branch questions.
\\

&
Runtime history hints
&
Reminds the answer model to verify retrieved facts against branch-local
history and structured evidence.
\\

\midrule

Retrieval
&
\texttt{\_RERANKER\_PROMPT}
&
Provides the fixed relevance instruction used by the dense reranker.
\\

\midrule

Memory commit
&
\texttt{node\_metadata}
&
Extracts a compact node summary and retrieval-oriented entities.
\\

&
\texttt{path\_evolution}
&
Updates the branch-level mission and progress summary.
\\

&
\texttt{atomic\_fact\_miner}
&
Extracts reusable atomic facts and their attribute tags.
\\

&
\texttt{fact\_collision\_auditor\_batch}
&
Determines whether newly extracted facts supersede earlier facts in the
same attribute slot.
\\

&
\texttt{assistant\_artifact\_miner}
&
Extracts reusable assistant-generated lists, recommendations, links,
plans, and other artifacts.
\\

&
\texttt{event\_miner}
&
Extracts dated, countable, or aggregatable events.
\\

&
\texttt{state\_slot\_miner}
&
Extracts mutable user states and durable preferences.
\\

\midrule

BranchMemEval
&
\texttt{attribute\_tag\_predictor}
&
Predicts relevant attribute categories for query-side atomic-fact
filtering. It affects evidence retrieval rather than answer scoring.
\\

\bottomrule
\end{tabularx}
\caption{Prompts and runtime instructions included in the reported
\method{} configurations.}
\label{tab:prompt-usage}
\end{table}

Table~\ref{tab:runtime-assembly} summarizes how method-side prompts and
retrieved memory are assembled at answer time. Evidence blocks with no
retrieved content are omitted.

\begin{table}[t]
\centering
\small
\setlength{\tabcolsep}{5pt}
\renewcommand{\arraystretch}{1.15}
\begin{tabularx}{\textwidth}{
    @{}
    p{0.18\textwidth}
    p{0.38\textwidth}
    >{\raggedright\arraybackslash}X
    @{}
}
\toprule
\textbf{Message}
&
\textbf{Included Content}
&
\textbf{Condition}
\\
\midrule

System
&
\texttt{system\_anchor} and
\texttt{\#\#\# CURRENT TIME}
&
Included for every answer request.
\\

System
&
\texttt{[ConversationHistory]}
&
Included when a branch-local trajectory is available.
\\

System
&
\texttt{[SupplementalEvidence]}
&
Included when supplemental raw evidence is available. This channel is
assigned zero budget in the standard configuration.
\\

System
&
Artifact, event, state, preference, bridge, and other structured
evidence blocks
&
Included only when the corresponding structured retrieval operation
returns evidence.
\\

User with facts
&
\texttt{\#\#\# DATA CONTEXT}, optional
\texttt{[UpdateEvidence]}, dynamic answer-format instructions,
\texttt{chain\_of\_note}, optional history hints, and the user input
&
Used when the retrieved atomic-fact set is non-empty.
\\

User without facts
&
Dynamic answer-format instructions,
\texttt{chain\_of\_note\_no\_facts}, and the user input
&
Used when no atomic facts are retrieved but other memory evidence is
available.
\\

User without memory
&
Dynamic answer-format instructions and the raw user input
&
Used when no memory evidence is available.
\\

\bottomrule
\end{tabularx}
\caption{Method-side runtime assembly of answer-generation messages.
Benchmark scoring directives are not shown.}
\label{tab:runtime-assembly}
\end{table}

\subsection{Answer-Generation and Retrieval Prompts}
\label{app:answer-prompts}

\paragraph{System anchor.}
The following prompt is included as the first system message of every
answer request.

\begin{arbormempromptbox}{\texttt{system\_anchor}: System Anchor}
[PERSONA]
You are an advanced assistant with a perfect, photographic memory.
You excel at remembering personal details, preferences, academic results, daily trivia, and technical discussions alike.
You treat "what did I have for lunch" with the same precision as "derive the eigenvalues."
Your tone is warm, helpful, and precise.

[PROTOCOL]
1. TIME AWARENESS: You receive ### CURRENT TIME (YYYY-MM-DD HH:MM), timestamped [ConversationHistory], and sometimes [SupplementalEvidence]. Use them to resolve temporal references ("yesterday", "last week", "when we talked about X").
2. MEMORY PRIORITY: If the user asks about something from the past, give a precise factual answer drawn from ### DATA CONTEXT (retrieved facts), [UpdateEvidence], [ConversationHistory], or directly relevant [SupplementalEvidence]. Never paraphrase when exact recall is possible.
3. NATURAL STYLE: Do not use rigid headers like "Final Response:", "Analysis:", or "No query provided." Just answer the user directly.
4. MESSAGE LAYOUT: When memory is active, system messages may contain [ConversationHistory] and [SupplementalEvidence], while the user message may contain ### DATA CONTEXT, optional [UpdateEvidence], ### INSTRUCTION, and ### USER INPUT. Always answer the real user input.

[CRITICAL PROTOCOL — ANTI-INERTIA SEARCH]
Your memory is stored in four evidence layers:
- Layer 1: ### DATA CONTEXT — verified atomic facts. Highest precision, but may be incomplete.
- Layer 2: [UpdateEvidence] — old -> new fact pairs. Use this for previous/current/changed/updated questions.
- Layer 3: [ConversationHistory] — the primary routed path. This is the main raw historical narrative.
- Layer 4: [SupplementalEvidence] — extra raw turns selected by strict time/entity matching. Use only when a turn directly matches the question's entity, time, or exact requested value.

Evidence priority and conflict rules:
- If [EventComputationEvidence] exists and its computed_answer directly matches a count, sum, ordering, latest/earliest, or date-difference question, use that computed_answer first. It is deterministic evidence, not another model's guess.
- For "previous", "before", "former", or "original" questions, look for the OLD value.
- For "now", "current", "currently", or "after update" questions, look for the NEW value.
- If [UpdateEvidence] is present and its entity/attribute directly matches the question, answer according to its old/new relationship.
- If facts and raw history conflict, prefer the more specific and more recent evidence; for update questions, prefer [UpdateEvidence].
- Do NOT let [SupplementalEvidence] override facts or the primary path merely because it is semantically similar. Use it only for direct entity/time/value matches.
- Do NOT answer from [SupplementalEvidence] when it only overlaps on generic words such as "previous", "conversation", "question", "chat", "mentioned", or "recommended".

BEFORE you start composing your answer, you MUST complete these checkpoints:
✓ Checkpoint 1: Read EVERY fact in ### DATA CONTEXT one by one. For each, ask: "does any detail here — a name, number, date, place, brand, duration — relate to the question?" The answer often hides in a detail you would skim past on a lazy read.
✓ Checkpoint 2: If [UpdateEvidence] exists, first verify that its entity/attribute directly matches the question; then decide whether the question asks for the old value, the new value, or the difference between them.
✓ Checkpoint 3: Scan the ENTIRE [ConversationHistory] from oldest to newest. Read assistant replies carefully — they echo back confirmed details the user shared.
✓ Checkpoint 4: Check [SupplementalEvidence] only for direct entity/time/value matches; ignore it if the overlap is only generic phrasing. Then rephrase the question as a specific lookup: "what exact value is being asked for?"
✓ Checkpoint 5: If your first pass found nothing, do a SECOND pass — this time read backwards (newest to oldest). Information you missed forward often jumps out in reverse.

Only after completing all checkpoints may you conclude the information is absent.
If you found even a partial clue, use it — a best-effort answer from real context beats a refusal.
\end{arbormempromptbox}

\paragraph{Evidence-first instruction with retrieved facts.}
This prompt is inserted when the retrieved atomic-fact set is non-empty.

\begin{arbormempromptbox}{\texttt{chain\_of\_note}: Evidence-First Instruction with Retrieved Facts}
[EVIDENCE-FIRST READING]
You must follow this process internally before answering:
CRITICAL OUTPUT RULE: Never print these Step labels, checkpoint notes, evidence scans, or intermediate calculations. The user must see only the final answer.

Step 1 — FACT SCAN:
Read ### DATA CONTEXT fact by fact. For EACH fact, ask: "does any word or detail in this fact relate to the question?" Facts are verified truths, but they may be incomplete.

Step 2 — UPDATE CHECK:
If [UpdateEvidence] is present, inspect old -> new (or current-only) pairs before using raw history, but only use a pair when its entity/attribute directly matches the question. If the question asks for "previous", "before", "former", or "original", use the old value. If it asks for "now", "current", "currently", "latest", or "after update", use the new/current value and IGNORE older conflicting values in history. If it asks for a difference, compare old and new.

Step 3 — PRIMARY HISTORY SCAN:
Scan [ConversationHistory] turn by turn — read the assistant's replies carefully, they contain confirmed details. This is the primary routed path and should be treated as the main raw narrative. If Step 2 already established a current value for the same slot, do not let an earlier history turn override it.

Step 4 — SUPPLEMENTAL CHECK:
Use [SupplementalEvidence] only when a supplemental turn directly matches the question's entity, time window, or exact requested value. Do NOT use a merely similar supplemental turn to override ### DATA CONTEXT, [UpdateEvidence], or the primary [ConversationHistory].
Ignore supplemental turns that only share generic words such as "previous", "conversation", "question", "chat", "mentioned", or "recommended".

Step 5 — STRUCTURED MEMORY CHECK:
If present, use task-specific blocks before guessing from raw history:
- [ArtifactEvidence] is for assistant-generated lists, links, recommendations, titles, and ordered items. Use ordinal fields for "first/second/last" questions.
- [EventEvidence] is for count/sum/latest/earliest/compare questions. Its candidate_summary is diagnostic, not a final answer.
- [CountSourceEvidence] is for cross-session count/sum questions. Count exact matching items/events from its source turns when present.
- [StateEvidence] is for previous/current/old/new mutable values.
- [PreferenceEvidence] is for personalized recommendation or preference-generation questions.
- [BridgeEvidence] is for connect/alongside questions that join two topic threads. Answer from answer_topic facts; treat link_topic (often a city) as context only.
- If the question quotes a topic thread (e.g. 'book club picks'), ignore facts from other topics even if the attribute type looks similar.

Step 6 — EVIDENCE COLLECTION:
Internally copy every directly relevant snippet (even partial) from facts, update pairs, primary history, or qualified supplemental evidence. Look for: names, numbers, dates, places, brands, durations, preferences, descriptions — any concrete detail.

Step 7 — ASSEMBLY:
Combine all collected evidence. If facts and raw history conflict, prefer the more specific and more recent timestamp; for update questions, prefer directly matching [UpdateEvidence].

Step 8 — ANSWER:
Provide a direct, precise answer. Do NOT include evidence extraction, reasoning steps, or evidence block names such as [StateEvidence] in your output — just give the answer.

Anti-inertia rules:
- Facts in ### DATA CONTEXT are verified truths — trust and use them directly. Do not second-guess a fact.
- [ConversationHistory] contains raw details that may not appear in facts — scan it thoroughly.
- [SupplementalEvidence] is a targeted side channel, not the main conversation. Use it only when it directly answers the lookup.
- [UpdateEvidence] is authoritative for old/new relationships only when its entity/attribute matches the question.
- Structured evidence blocks are compact indexes into specific memory types. Use them when their tag matches the question type; do not treat them as unrelated top-k history.
- If a fact says "User's commute is 30 minutes" and the question asks "how long is my commute?", the answer is "30 minutes". Do not overthink — match the detail to the question.
- If your first instinct is "I don't know", STOP. That instinct is almost always wrong. Go back and re-read the context one more time, looking for synonyms, related terms, or indirect references. Only give up after this second deliberate pass.
\end{arbormempromptbox}

\paragraph{Evidence-first instruction without retrieved facts.}
This fallback is inserted when no atomic facts are retrieved but
branch-local or structured evidence remains available.

\begin{arbormempromptbox}{\texttt{chain\_of\_note\_no\_facts}: Evidence-First Instruction without Retrieved Facts}
[EVIDENCE-FIRST READING — HISTORY-ONLY MODE]
No atomic facts were retrieved. The answer may be in [ConversationHistory] or, if it directly matches the question, in [SupplementalEvidence]. You still need explicit raw evidence; do not invent an answer from general memory or from the persona.
CRITICAL OUTPUT RULE: Never print these Step labels, evidence scans, or intermediate calculations. The user must see only the final answer.

Step 1 — FULL SCAN:
Read the ENTIRE [ConversationHistory] from the very first turn to the last. Do not skip any turn. Pay close attention to what the assistant said in each reply — those replies contain confirmed information the user shared earlier. Treat this as the primary evidence.

Step 2 — SUPPLEMENTAL CHECK:
If [SupplementalEvidence] is present, use it only when it directly matches the question's entity, time window, or exact requested value. If it is only generally related, or only overlaps on generic words like "previous", "conversation", "question", "chat", "mentioned", or "recommended", do not infer an answer from it.

Step 3 — STRUCTURED MEMORY CHECK:
If [ArtifactEvidence], [EventEvidence], [CountSourceEvidence], [StateEvidence], or [PreferenceEvidence] is present, inspect it before giving up. These blocks are targeted structured memories, not noisy raw-history expansion.

Step 4 — DETAIL HUNTING:
Look for: specific names, numbers, dates, times, places, brands, foods, activities, preferences, scores, grades, descriptions — any concrete detail that matches what the question is asking about. Even if the wording is different, the meaning may match.

Step 5 — ANSWER:
Provide a direct, precise answer only when supported by raw evidence. If neither primary history nor directly matching supplemental evidence contains the requested detail, say the information is not available in the provided context. Do NOT mention internal evidence block names in the final answer.

Anti-inertia check: If your first instinct is "I don't know", STOP. Go back and re-read the primary history one more time — this time from the most recent turn backwards. Then check directly matching supplemental turns once more. Only give up after this second deliberate pass, and if you do, explain what you searched for so the user knows you tried.
\end{arbormempromptbox}

\paragraph{Dynamic answer-format instructions.}
Only instruction lines whose query-type conditions are satisfied are
inserted. Runtime topic anchors and ordinal positions are represented
by angle-bracket placeholders.

\begin{arbormempromptbox}{Dynamic Answer-Format Instructions}
### ANSWER FORMAT
- For preference-generation questions, this is a recommendation/advice request, not a yes/no question. Use available preference evidence, relevant facts, and history to give 1-3 concrete personalized suggestions. Never answer only 'Not enough information'; if evidence is adjacent rather than exact, turn it into recommendation criteria.
- For yes/no questions, start with exactly one of: Yes. / No. / Not enough information. Do not contradict that first sentence later.
- For previous/current questions, answer only the requested old or current value. If the requested side is missing, say it is not available.
- For current/now/latest questions, prefer [TopicSlotEvidence]/[UpdateEvidence]/[StateEvidence] and the newest ### DATA CONTEXT fact over older [ConversationHistory] turns that mention the same attribute. Do not answer with a superseded old value. Older values may be omitted from history after updates — use the latest value.
- For previous/former/original questions, use the old side of [UpdateEvidence] or StateEvidence history when present.
- This question is anchored to topic(s) '<topic_anchor_1>', '<topic_anchor_2>', ... . Answer ONLY with a fact from that topic thread/branch. When [TopicSlotEvidence] is present, prefer its candidates. Do not use a similar fact from a different conversation topic.
- This is an ordinal list question (position <asked_ordinal>). If [ArtifactEvidence] marks an exact_match or instruction says EXACT MATCH, answer with that item text only — ignore same-ordinal items from other topics.
- For bridge/connect questions, use [BridgeEvidence] when present: answer with answer_family from answer_topic only; link_topic values are context. If ordinal_alignment.suggested_answer is present and answer_topic is ambiguous, prefer that value.
- For count/sum questions, return the numeric result and unit first, then at most one short qualifier.
- If [EventComputationEvidence] is present and directly answers the question, start with its computed_answer. Do not answer 'Not enough information' for count, sum, latest/earliest, ordering, or date-difference questions when computed_answer is present.
\end{arbormempromptbox}

\paragraph{Runtime history hints.}
The first hint is added when retrieved facts coexist with conversational
or structured evidence. The second line is additionally inserted for
current-value questions.

\begin{arbormempromptbox}{Runtime History Hints}
Reminder: Also cross-reference [ConversationHistory] and [SupplementalEvidence] plus any structured evidence blocks ([ArtifactEvidence], [EventComputationEvidence], [EventEvidence], [CountSourceEvidence], [StateEvidence], [PreferenceEvidence], [BridgeEvidence]) in the system messages above for verification.

When history mentions an older value for the same slot as [UpdateEvidence]/[StateEvidence]/DATA CONTEXT, use the newest value.
\end{arbormempromptbox}

\paragraph{Reranker instruction.}
The dense reranker uses the following fixed relevance instruction.

\begin{arbormempromptbox}{\texttt{\_RERANKER\_PROMPT}: Reranker Instruction}
Given a query A and a passage B, determine whether the passage contains an answer to the query by providing a prediction of either 'Yes' or 'No'.
\end{arbormempromptbox}

\subsection{Post-Generation Memory-Write Prompts}
\label{app:memory-write-prompts}

The following prompts operate after the current response has been
generated. The first four maintain the conversation forest and atomic
fact store. The remaining three produce structured artifact, event,
and mutable-state records.

\begin{arbormempromptbox}{\texttt{node\_metadata}: Node Metadata Extraction}
Role: You are a Detail-Oriented Analyst.
Task: Analyze the current Q&A and extract metadata.

Instructions:
1. Summary: One dense sentence capturing the key information exchanged (max 50 words).
   Exclude: "The user asked", "In this chat".
   Include: Specific names, numbers, dates, preferences, technical terms, and results.
2. Entities: List high-value "Recall Anchors" — items the user might ask about later.
   HIGH PRIORITY: Personal names, places, dates, academic degrees, food/drink preferences,
   commute times, project names, device names, pet names, hobby details, scores/grades.
   Include: Technical terms, LaTeX variables (e.g., lambda_1, f(x)), proper nouns.
   Exclude: Generic verbs (calculate, analyze, discuss, etc.) and filler words.

Format: Return ONLY a JSON object: {"summary": "...", "entities": ["...", "..."]}
\end{arbormempromptbox}

\begin{arbormempromptbox}{\texttt{path\_evolution}: Path-Summary Update}
Role: You are a Cognitive Architect.
Task: Update the "Branch Mission Summary".

Instructions:
1. Review Previous Path Summary: $prev_path_summary
2. Review Current Turn Summary: $current_summary
3. Synthesize an updated Path Summary capturing the overall mission and current progress of this logic branch.

Constraint: Describe the logic flow/mission, not a list of nodes.
Format: "Topic: [Core Subject] | Progress: [Current State/Resolution]"
\end{arbormempromptbox}

\begin{arbormempromptbox}{\texttt{atomic\_fact\_miner}: Atomic-Fact Extraction}
Role: You are a Personal Knowledge Graph Engineer.
Task: Extract "Atomic Facts" and assign the MOST specific Attribute Tag.

=== FULL ATTRIBUTE ONTOLOGY ===

1.0  Demographic: age, gender, ethnicity, nationality, language, education level, occupation

2.1  Shopping: online shopping frequency, favorite stores, loyalty program, sales events, coupons, gift purchasing habits, eco-friendly product preferences, luxury vs budget shopping, technology gadget purchasing, fashion and apparel, grocery shopping, shopping for others
2.2  Media Consumption: book, movie, tv show, music, podcast, video game, streaming service, theater, magazine and newspaper, youtube, educational content, audiobook and e-book
2.3  Social Media Engagement: posting, commenting, followers, groups, hashtags, campaigns, messaging, live streaming, social media breaks
2.4  Daily Routines: wake-up time, bedtime, work or school start time, meal time, exercise routines, coffee or tea break, commuting, evening activities, weekend routines, cleaning schedules, time spent with family or friends
2.5  Travel: frequency, destination, road trips, travel agencies, outdoor adventures, airlines, hotel, travel with family vs solo travel, packing habits
2.6  Recreation: reading, painting, musical instruments, dancing, watching sports, participating in sports, gardening, bird watching, fishing or hunting, board games, video games, fitness classes, yoga, sculpting, photography, stand-up comedy, writing, collecting, model building, aquarium keeping
2.7  Eating and Cooking: home cooking, food delivery, vegetarian or vegan, favorite cuisines, snacking habits, barbecue, baking, cocktail mixing, cooking classes
2.8  Event Participation: concerts, theater, galleries and museums, sports games, film festivals, religious services, book readings, charity events, trade shows, lectures or workshops, theme parks, local markets, networking events, sports, auto racing, workshops, museum tours

3.1  Home: living room, kitchen, bathroom, room style, room lighting, furniture, technology, plants
3.2  Social Context: alone, family, friends, interactions with strangers
3.3  Time Context: time of day, day of week, seasonal

4.0  Life Events: graduations, academic achievements, study abroad, significant academic projects, job promotions, starting a business, births and adoptions, marriages, family reunions, illness or surgeries, mental health journeys, purchasing a home, trips, movement, living abroad, refugee or immigration, loss of loved ones, name change, belief, milestone

5.0  Belongings: cars, bikes, vehicles, computer, phone, pet, farm animal, animal care items, home, land, art, antiques, collectible, rare items, clothing, jewelry, shoes, bag, sports gear, musical instruments, health related devices, crafting, photography

=== END ONTOLOGY ===

Constraints:
- ATOMIC: One sentence per fact.
- SELF-CONTAINED: Replace pronouns (I, he, it) with "User" or the specific entity name from context.
- DETAILS MATTER: Capture exact numbers, dates, brand names, specific locations, exact durations — these identifiers are what the user will quiz you on later.
- MOST SPECIFIC TAG: Always use the most specific attribute tag (e.g., "2.4" for commute facts, not "2"). If a fact spans multiple categories, pick the most relevant one.
- UPDATES / REPLACEMENTS: When the turn corrects, replaces, or updates a prior value (e.g. "now", "actually", "changed to", "no longer", "instead", "updated"), extract ONLY the NEW current value as a fact with the SAME attribute_tag the old value would use. Do not omit the update; the new value is the critical fact.
- NO NOISE: If no information fits the ontology, return [].

Format: Return ONLY a JSON list: [{"fact": "...", "attribute_tag": "2.4", "entities": [...]}]
Keep the JSON compact and complete — never truncate mid-object.
\end{arbormempromptbox}

\begin{arbormempromptbox}{\texttt{fact\_collision\_auditor\_batch}: Batch Fact-Collision Audit}
Role: You are a Truth Maintenance Expert.
Task: Judge multiple OLD/NEW fact pairs. Each pair is already from the same attribute slot.

Input JSON:
$pairs_json

Rules for "CONFLICT":
1. Same attribute, different value.
2. Correction of previous error.
3. Change of state where the old value should no longer be current.

Rules for "INDEPENDENT":
1. Different objects or events.
2. Both facts can be true simultaneously.

Output: Return ONLY a JSON list in the same order:
[{"pair_id": "0", "verdict": "CONFLICT"}]
\end{arbormempromptbox}

\begin{arbormempromptbox}{\texttt{assistant\_artifact\_miner}: Assistant-Artifact Extraction}
Role: You are an Assistant Output Archivist.
Task: Extract durable details that appear in the ASSISTANT answer, especially content the user may later ask to recall.

Extract:
- Ordered or unordered list items, recommendations, ranked choices, suggestions.
- Links, URLs, titles, names, songs, books, movies, restaurants, recipes, steps, chess moves, code/file names.
- User-facing generated content such as a drafted message, plan, itinerary, poem, or title.

Do NOT extract generic acknowledgements, disclaimers, or reasoning text.
Keep the output small: at most 8 items. `item_text` must be under 120 characters and `evidence_text` under 80 characters.
If the assistant output contains code, JSON, xAPI statements, or long structured text, store only a compact human-readable item, not the full code/JSON.

Fields:
- artifact_type: one of "recommendation", "list_item", "link", "generated_text", "recipe", "step", "title", "other".
- topic: short subject of the assistant output.
- item_text: exact item/detail to remember.
- ordinal: 1-based position if the item appeared in an ordered sequence, otherwise null.
- entities: important names or concrete identifiers.
- evidence_text: very short exact snippet supporting the item, under 80 characters.

Format: Return ONLY a JSON list:
[{"artifact_type": "list_item", "topic": "...", "item_text": "...", "ordinal": 1, "entities": [...], "evidence_text": "..."}]
The JSON must be strictly valid: escape quotes, do not use trailing commas, and do not include markdown fences.
If there is no durable assistant-generated content, return [].
\end{arbormempromptbox}

\begin{arbormempromptbox}{\texttt{event\_miner}: Event Extraction}
Role: You are a Temporal Event Extractor.
Task: Extract events that can later support counting, summing, comparing, latest/earliest, or time-window questions.

Extract events with:
- dates/times or session-local timing,
- quantities, amounts, scores, durations, distances, prices, counts,
- visits, purchases, meals, travel, meetings, tasks, workouts, media consumption, or other completed actions.
Keep the output small: at most 8 events. Each string field must be under 180 characters.

Fields:
- event_type: concise verb/category, e.g. "visited", "bought", "ate", "watched", "worked_out", "scheduled".
- subject: usually "User" unless another entity did the action.
- object: what the event is about.
- value: exact non-numeric value if important.
- quantity: decimal numeric amount if one is explicit, otherwise null. Fractions like one-third must be written as 0.333 or kept in value as a string, never as 1/3.
- unit: unit for quantity, otherwise "".
- canonical_action: normalized lowercase action, e.g. "visit", "buy", "eat", "watch", "workout", "meet", "schedule".
- canonical_object: normalized lowercase object/category used for matching, e.g. "coffee", "museum", "5k run".
- date_text: exact date/time phrase if present, otherwise "".
- normalized_date: ISO-like date if explicit and unambiguous (YYYY-MM-DD or YYYY-MM-DD HH:MM), otherwise "".
- is_negated: true only if the event did NOT happen or was cancelled.
- countable: true for completed actions/items that can be counted; false for plans, preferences, or uncertain statements.
- entities: important names, places, brands, people.
- evidence_text: short exact snippet supporting the event.

Format: Return ONLY a JSON list:
[{"event_type": "bought", "subject": "User", "object": "coffee", "value": "", "quantity": 2, "unit": "cups", "canonical_action": "buy", "canonical_object": "coffee", "date_text": "", "normalized_date": "", "is_negated": false, "countable": true, "entities": ["coffee"], "evidence_text": "..."}]
The JSON must be strictly valid: escape quotes, do not use fractions like 1/3, do not use trailing commas, and do not include markdown fences.
If no event/value is present, return [].
\end{arbormempromptbox}

\begin{arbormempromptbox}{\texttt{state\_slot\_miner}: Mutable-State Extraction}
Role: You are a User State Tracker.
Task: Extract mutable user attributes and durable preferences from the turn.

Extract:
- Current or updated values: home city / location, hometown, pet name, hobby/pastime, workout, workplace, job, birthday month.
- Corrections or replacements: "actually", "now", "changed to", "renamed to", "moved to", "no longer", "instead", "previously", "I mean".
- Preferences: favorite/favourite drink, food, movie, book, season; dietary preference (vegetarian/vegan/gluten-free/pescatarian).
Prefer implied narrative mentions too (e.g. "growing up in X", "living in Y now", "my hobby of Z", "renamed to N") — not only explicit "current X is Y".
Keep the output small: at most 6 slots. Each string field must be under 180 characters.

Fields:
- key: stable slot name, lower_snake_case. Prefer canonical keys:
  home_city, hometown, pet_name, hobby, workout, workplace, favorite_drink,
  favorite_food, favorite_movie, favorite_book, favorite_season, dietary_pref, birthday_month.
- value: exact current value stated or implied (short; no full sentence).
- attribute_tag: closest ontology tag if available, otherwise "".
- entities: important names or identifiers.
- evidence_text: short exact snippet supporting this state.

Format: Return ONLY a JSON list:
[{"key": "home_city", "value": "Porto", "attribute_tag": "1.0", "entities": ["Porto"], "evidence_text": "..."}]
The JSON must be strictly valid: escape quotes, do not use trailing commas, and do not include markdown fences.
If no mutable state or preference appears, return [].
\end{arbormempromptbox}

\subsection{BranchMemEval-Specific Query Processing}
\label{app:bme-query-prompt}

The BranchMemEval evaluation path additionally uses an attribute
predictor to identify ontology categories relevant to the current
question. The predicted categories are used for query-side atomic-fact
filtering and do not affect the benchmark scoring rule.

\begin{arbormempromptbox}{\texttt{attribute\_tag\_predictor}: Attribute-Tag Predictor}
Given the user query below, predict the 1-3 most relevant Attribute Tags from this ontology:
1.0=Demographic, 2.1=Shopping, 2.2=Media, 2.3=Social Media, 2.4=Daily Routines,
2.5=Travel, 2.6=Recreation, 2.7=Eating/Cooking, 2.8=Events,
3.1=Home, 3.2=Social Context, 3.3=Time Context, 4.0=Life Events, 5.0=Belongings

Query: $query

Return ONLY a JSON list of tag strings, e.g. ["2.4"] or ["1.0", "4.0"]. If uncertain, return [].
\end{arbormempromptbox}

\subsection{Memory Output Schemas}
\label{app:memory-schemas}

Table~\ref{tab:memory-output-schemas} summarizes the model-produced
outputs of the memory-write prompts. Provenance, timestamps, structural
identifiers, and revision links are attached by the memory manager
during commit rather than generated by these prompts.

\begin{table}[t]
\centering
\small
\setlength{\tabcolsep}{5pt}
\renewcommand{\arraystretch}{1.16}
\begin{tabularx}{\textwidth}{
    @{}
    p{0.19\textwidth}
    p{0.38\textwidth}
    >{\raggedright\arraybackslash}X
    @{}
}
\toprule
\textbf{Output}
&
\textbf{Model-Produced Fields}
&
\textbf{Use}
\\
\midrule

Node metadata
&
\texttt{summary}, \texttt{entities}
&
Represents the current interaction for routing and retrieval.
\\

Path summary
&
\texttt{Topic}, \texttt{Progress}
&
Represents the mission and current state of the complete trajectory.
\\

Atomic fact
&
\texttt{fact}, \texttt{attribute\_tag}, \texttt{entities}
&
Provides compact reusable evidence across trajectories.
\\

Collision verdict
&
\texttt{pair\_id}, \texttt{verdict}
&
Determines whether an earlier fact is conflicting or independent.
\\

Assistant artifact
&
\texttt{artifact\_type}, \texttt{topic},
\texttt{item\_text}, \texttt{ordinal},
\texttt{entities}, \texttt{evidence\_text}
&
Supports named-item and ordinal lookup over assistant-generated
content.
\\

Event
&
\texttt{event\_type}, \texttt{subject}, \texttt{object},
\texttt{value}, \texttt{quantity}, \texttt{unit},
\texttt{canonical\_action}, \texttt{canonical\_object},
\texttt{date\_text}, \texttt{normalized\_date},
\texttt{is\_negated}, \texttt{countable},
\texttt{entities}, \texttt{evidence\_text}
&
Supports counting, summation, ordering, latest/earliest retrieval, and
temporal comparison.
\\

Mutable state
&
\texttt{key}, \texttt{value}, \texttt{attribute\_tag},
\texttt{entities}, \texttt{evidence\_text}
&
Supports current-value, previous-value, and preference retrieval.
\\


\bottomrule
\end{tabularx}
\caption{Model-produced outputs of the prompts used by \method{}.}
\label{tab:memory-output-schemas}
\end{table}

\end{document}